\documentclass{article}

\usepackage{arxiv2027,times}
\usepackage{amsmath,amssymb,booktabs,graphicx,xcolor,array,tabularx,longtable}
\usepackage[normalem]{ulem}
\usepackage{subcaption}
\usepackage{enumitem,listings,hyperref,url,wrapfig,float}
\usepackage{amsmath}
\usepackage{amssymb}
\usepackage{algorithm}
\usepackage{algpseudocode}

\hypersetup{
  colorlinks=true,
  linkcolor=black,
  citecolor=black,
  urlcolor=black,
  pdftitle={COEVO: Co-Evolving Context and Parameters for Recursive Self-Improvement},
  pdfauthor={}
}

\definecolor{pending}{RGB}{180,35,40}
\newcommand{\tbd}[1]{\textcolor{pending}{\textbf{[#1]}}}
\newcommand{\pendingtext}[1]{{\color{pending}#1}}
\newcommand{\method}{COEVO}

\newcolumntype{Y}{>{\raggedright\arraybackslash}X}
\title{COEVO: Co-Evolving Context and Parameters for Recursive Self-Improvement}
\shorttitle{COEVO: Co-Evolving Context and Parameters for Recursive Self-Improvement}
\author{
\textbf{Siwei Chen}$^{1}$ \quad
\textbf{Xinping Bao}$^{1}$ \quad
\textbf{Xinyu Cai}$^{1}$ \quad
\textbf{Yuan Cao}$^{1}$ \quad
\textbf{Wan Jiang}$^{1}$ \quad
\textbf{Shaohong Chen}$^{2}$ \\
$^{1}$Peking University \qquad
$^{2}$Emotional Machine
}

\begin{document}
\vspace*{-0.8cm}
\maketitle
\vspace*{-0.4cm}
\begin{abstract}

Recursive self-improvement (RSI) seeks to move large language models beyond static training pipelines toward systems that can participate in improving their own future behavior. Existing approaches largely follow two directions: updating model parameters through online learning, or improving the external context through search, reflection, and prompt optimization. Although both mechanisms can support continued improvement, they are typically studied independently. This separation overlooks an important interaction: the context shapes the experience from which a model learns, while an evolving model may interpret and utilize the same context differently over time. We therefore formulate RSI as a problem of parameter--context co-evolution, where model parameters and the learning context adapt within a shared feedback loop. We introduce \method{}, a framework that updates model parameters from on-policy experience while adapting contextual guidance according to the state of the evolving policy. Policy entropy and prompt-conditioned attention are used as complementary signals to guide this adaptation. Experiments show that \method{} consistently improves task performance over fixed-context reinforcement learning and produces policies that are more robust to changes in system prompts. More broadly, our results suggest that external context should be viewed not merely as a fixed interface to a large language model, but as an adaptive component of recursive self-improvement.

\end{abstract}

\section{Introduction}

Large language models (LLMs) have progressed from systems whose capabilities are primarily determined by large-scale pretraining to increasingly adaptive systems whose behavior can be further improved through post-training and interaction with external feedback. In particular, on-policy reinforcement learning allows a model to generate its own trajectories, receive task-level feedback, and update its parameters from the resulting experience~\citep{shao2024deepseekmath,yu2025dapo,feng2025retool}. This transition points toward a broader objective beyond conventional model training: recursive self-improvement (RSI), in which the current model actively participates in constructing the process that improves its future versions. Rather than treating learning as a sequence of externally specified optimization steps, RSI seeks to close the feedback loop between the state of a model, the experience it generates, and the mechanisms used to produce subsequent improvements.

Existing approaches to RSI in LLMs can be understood through two complementary forms. The first operates on the model itself, using online learning or reinforcement learning to continually update model parameters. Parameter adaptation can internalize new reasoning strategies, but the learning procedure typically assumes that the training context remain fixed throughout optimization ~\citep{shinn2023reflexion,yang2023opro,guo2023evoprompt}. The second operates on the model's external context: methods based on reflection, prompt search, evolutionary optimization, or linguistic memory iteratively modify intermediate scaffolds without directly changing model parameters~\citep{fernando2023promptbreeder,zelikman2023stop,agrawal2025gepa,zhang2025dgm}. Such approaches can rapidly alter how existing capabilities are elicited, but they generally optimize the context for a model whose underlying policy is treated as fixed during optimization.

However, parameter learning and context optimization are intrinsically coupled: the context shapes the trajectories used for parameter updates, while parameter updates change how that context is interpreted and utilized. Context effectiveness is therefore policy dependent, making RSI naturally a co-evolution problem. Although ESPL~\citep{zhang2026espl} jointly evolves prompts and model parameters, its prompt evolution remains primarily driven by downstream reward, effectively treating context adaptation as a black-box outer-loop search. Such a signal can rank candidate prompts, but provides little information about why a context is ineffective or how it should change as the policy evolves. We instead make context evolution explicitly policy-aware, using the policy's exploration state and context utilization to guide targeted adaptation.

\begin{figure}[t]
\centering
\includegraphics[width=\linewidth]{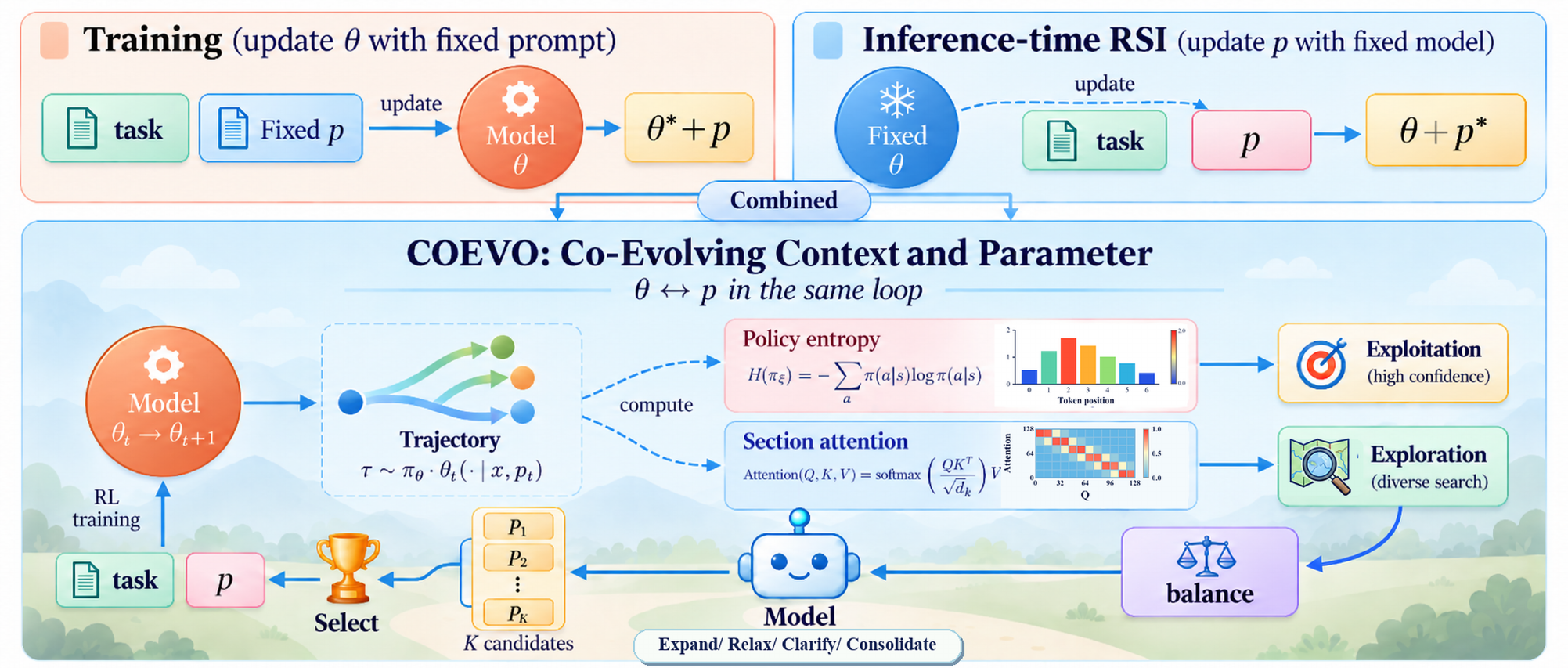}
\caption{The context--parameter feedback loop in \method.}
\label{fig:loop}
\vspace{-18pt}
\end{figure}

We propose \method, a reinforcement-learning-based framework for parameter--context co-evolution that closes the feedback loop between policy learning and context adaptation. During training, the current policy generates on-policy experience to update model parameters, while its evolving state guides subsequent context refinement. Specifically, we characterize the evolving policy using policy entropy, which reflects the model's exploration state, and prompt-conditioned attention, which measures its utilization of contextual guidance. These signals enable the context to adapt to the changing policy rather than being optimized solely on reward. The updated context then shapes future on-policy experience, forming a recursive interaction between parameter learning and context evolution. Figure~\ref{fig:loop} provides a conceptual illustration of \method. In our evaluated settings, \method\ achieves stronger task performance than fixed-prompt RL baselines. Ablation studies also show that the policy-aware signals improve over a reward-only prompt-evolution variant inspired by ESPL, indicating that entropy and attention signals benefit RL training and provide effective guidance for context evolution.

Overall, this paper makes the following contributions:
\begin{itemize}[leftmargin=*]
\item We formulate LLM recursive self-improvement as a general and unified parameter--context co-evolution problem, where model updates change how contextual guidance is interpreted, while context updates reshape the on-policy experience used for subsequent learning.

\item We introduce \method, a bounded RSI framework that jointly evolves model parameters and system context within a unified reinforcement-learning loop. Using policy entropy and prompt-conditioned attention to characterize the current policy state, \method\ performs targeted context adaptation and improves over standard RL algorithms.

\item We show that exploration and context utilization provide complementary signals for context evolution: entropy reveals changes in behavioral diversity, while attention identifies how the strategy scaffold is being used. Ablations demonstrate that removing either signal leads to distinct failure modes, highlighting the importance of policy-aware context adaptation.

\end{itemize}
\vspace{-6pt}
\section{Problem Formulation and Design Hypotheses}
\label{sec:observations}
\subsection{Context affects the learning distribution}
Training context determines which experiences are available for learning. Given policy parameters $\theta_t$ and context $p_t$, an on-policy trajectory $\tau_t$ is generated by rolling out the policy $\pi_{\theta_t}(\cdot\mid x,p_t)$ for task $x$ under context $p_t$, and this trajectory then contributes to the policy-gradient estimate. The dependence is thus given by:
\[
p_t\ \longrightarrow\ \tau_t\sim\pi_{\theta_t}(\cdot\mid x,p_t)
\ \longrightarrow\ \widehat{\nabla_\theta J}(\theta_t,p_t)
\ \longrightarrow\ \theta_{t+1}.
\]
Even with a fixed task reward and optimizer, revising $p_t$ changes the trajectory distribution entering the next update. This distinguishes training-context adaptation from choosing an inference wrapper after learning has finished.

\begin{figure}[t]
    \centering

    \begin{minipage}[t]{0.49\linewidth}
        \centering
        \includegraphics[width=\linewidth]{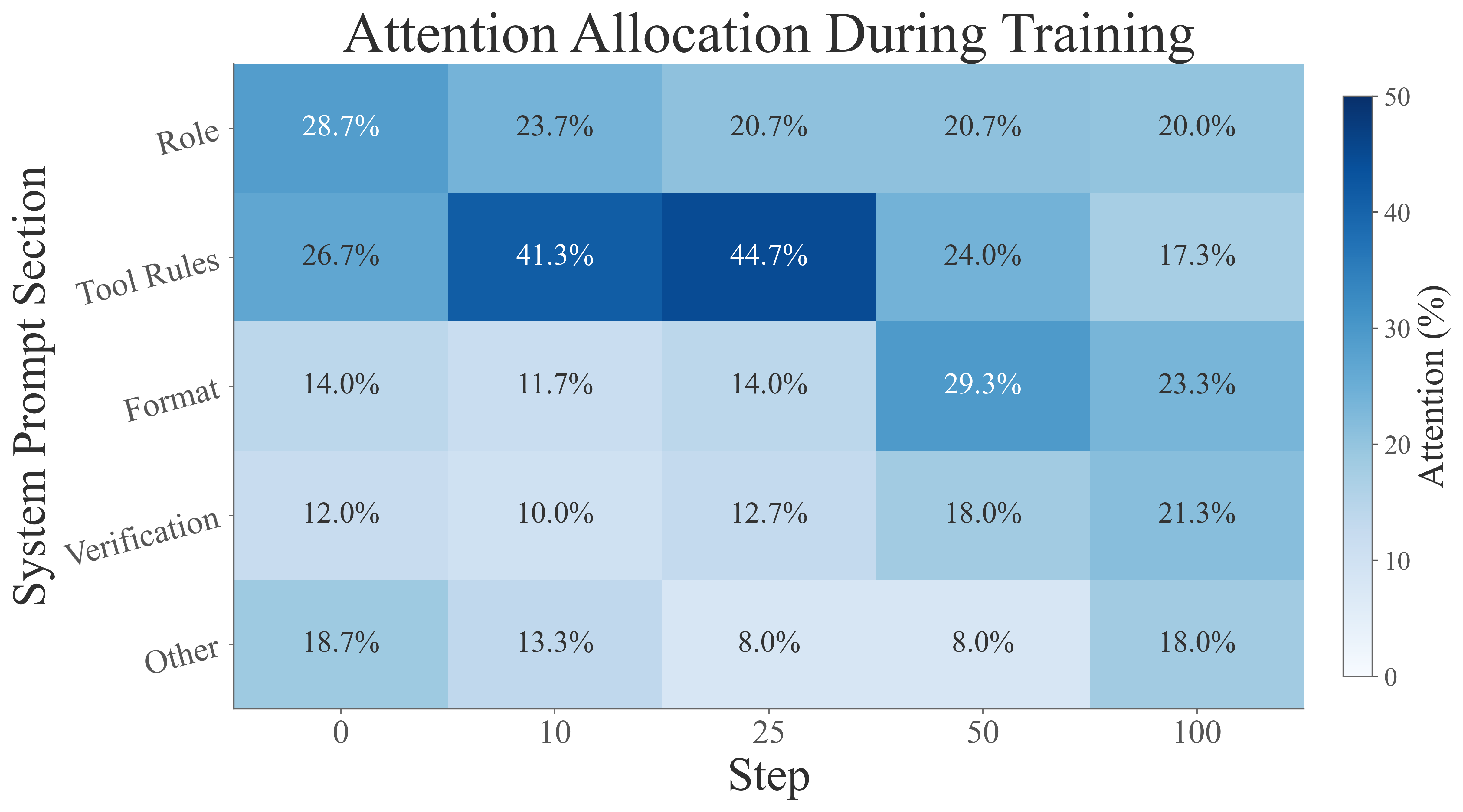}
        
        \vspace{-1mm}
        {\small \textbf{(a) Prompt-section attention allocation}}
    \end{minipage}
    \hfill
    \begin{minipage}[t]{0.49\linewidth}
        \centering
        \includegraphics[width=\linewidth]{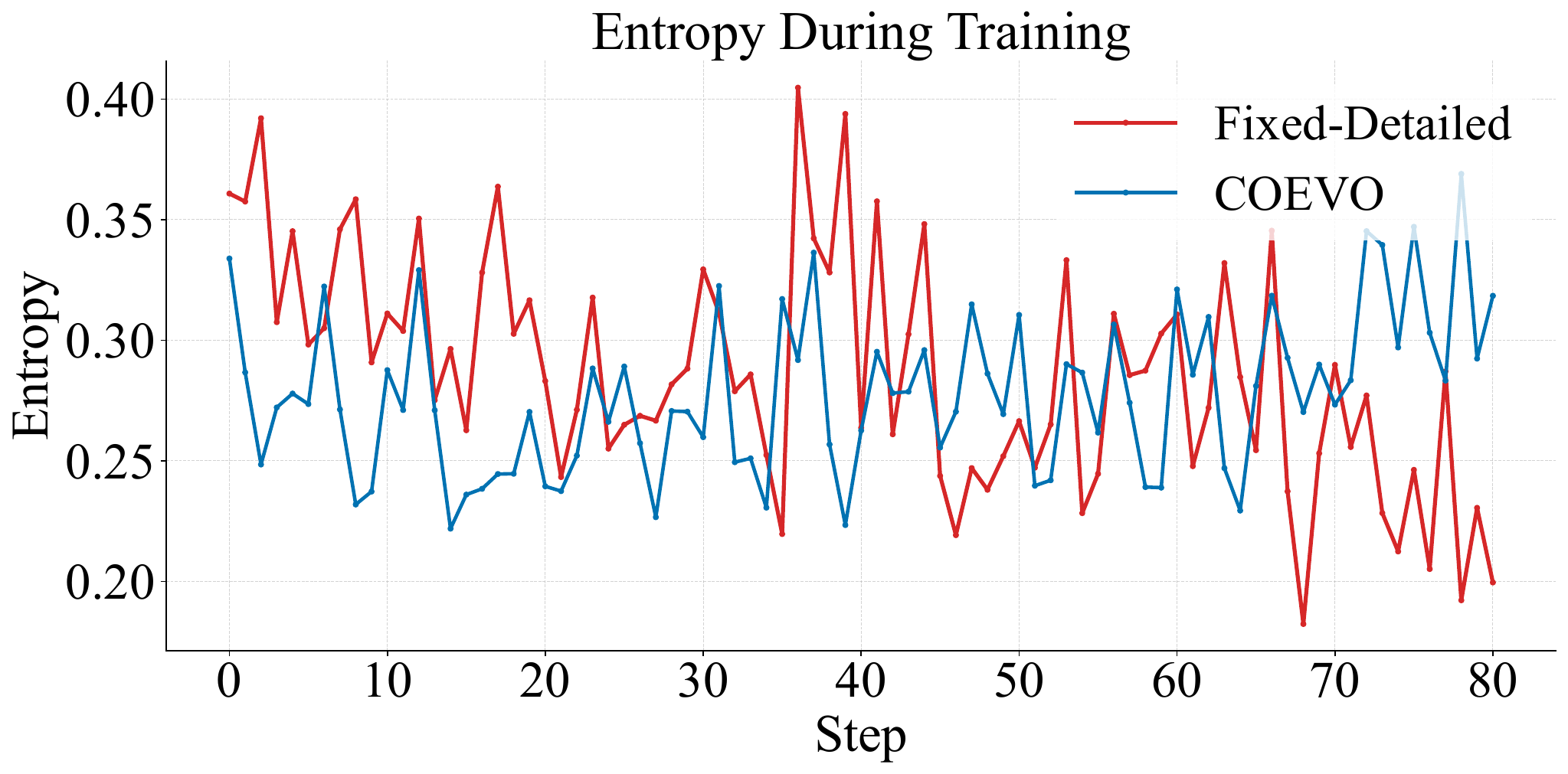}
        
        \vspace{-1mm}
        {\small \textbf{(b) Policy entropy}}
    \end{minipage}

    \vspace{-1mm}
    \caption{
        \textbf{Policy-state signals evolve throughout training.}
        \textbf{(a)} Relative attention allocation across functional prompt
        sections changes substantially over training, indicating that the
        policy's utilization of the same context is non-stationary.
        \textbf{(b)} Policy entropy captures the concurrent evolution of
        behavioral exploration: under a fixed detailed context, entropy
        progressively decreases, whereas \method{} maintains broader
        exploration through context adaptation.
        Together, these dynamics motivate using both context utilization
        and policy exploration to guide context evolution.
    }
    \label{fig:policy-state-dynamics}
    \vspace{-12pt}
\end{figure}

\subsection{Context Usefulness May Be Policy-Dependent}

The same scaffold can induce different behavior under different parameters.
Guidance that helps an early policy organize a solution may become redundant
or restrictive afterward, while an instruction ignored by one
policy may be actively used by another. We therefore hypothesize that a scaffold's usefulness depends on the current policy state, not on the prompt alone.

Figure~\ref{fig:policy-state-dynamics} provides an empirical illustration of
this non-stationarity. The attention heatmap (left) shows that attention is
redistributed across functional prompt sections as training progresses,
indicating that the policy's utilization of the same contextual scaffold
changes over time. Meanwhile, the entropy trajectories (right) reveal
concurrent changes in behavioral exploration: under a fixed detailed context,
the policy becomes progressively more concentrated, whereas adaptive context
evolution maintains broader exploration. Importantly, these observations do not imply a universal ordering of short,
detailed, or strongly attended prompts. Rather, they suggest that context
usefulness is relational: it depends on both how the current policy utilizes
the available guidance and how broadly the policy is exploring. This motivates
treating context revision as a state-dependent intervention, where the form of
an update is conditioned on the evolving policy state.

\subsection{Context revision as a state-dependent intervention}

Motivated by the dynamics in Figure~\ref{fig:policy-state-dynamics}, we characterize the current policy along two
complementary coordinates: policy entropy captures behavioral exploration, while section-level attention provides an operational proxy for context
utilization. Their combination helps distinguish policy states that may exhibit similar outcome feedback but call for different context interventions. Low exploration accompanied by weak use of a strategy section suggests a different intervention from low exploration accompanied by strong use. The same distinction applies when exploration is relatively high.

These coordinates guide candidate generation, while frozen-weight comparisons test candidate's immediate effects. Subsequent learning under shared evaluation contexts tests hypothesis that state-conditioned revision improves parameter acquisition. Attention remains an utilization proxy, and the reference band a calibration device; neither assumed to identify an optimal learning state.

\section{Co-Evolving Context and Parameters}
\label{sec:method}

\subsection{Parameter--Context Co-Evolution}
\label{sec:framework}

We formulate recursive self-improvement as a coupled learning process in which parameters and context evolve together. At iteration $t$, the state of the learning system is represented by $(\theta_t,p_t)$, where $\theta_t$ denotes the current model parameters and $p_t$ denotes the system prompt used to condition task interaction. We decompose the prompt as $p_t=c\oplus s_t$, where $c$ contains the fixed task specification and interface constraints, while $s_t$ is an editable strategy scaffold. For a task $x$ sampled from the training distribution and a trajectory $\tau$ generated by the current policy, the task objective is
\begin{equation}
J(\theta_t,p_t)
=
\mathrm{E}*{x,\tau}
\left[
R(x,\tau)
\right],
\qquad
\tau\sim\pi*{\theta_t}(\cdot\mid x,p_t).
\label{eq:objective}
\end{equation}
The reward remains the optimization objective throughout training; entropy and attention are used only to characterize the current policy state and guide context evolution.

Training alternates between parameter learning and context adaptation. During an RL window, the prompt $p_t$ is held fixed and the policy collects fresh on-policy trajectories for parameter updates. After the window, the updated checkpoint is frozen temporarily and used to determine whether the current strategy scaffold is still well matched to the policy. If a context revision is accepted, the resulting prompt $p_{t+1}$ is used to generate the next round of on-policy experience. Consequently, a context update affects not only how the current model behaves, but also which trajectories become training data for the next parameter update. Conversely, changes in $\theta_t$ alter how the model interprets and uses the same contextual guidance. The two components therefore form a feedback loop: context evolution changes future learning experience, while parameter learning changes the policy for which that context must be effective.

This formulation realizes a bounded form of RSI. The model parameters and editable strategy context participate in repeated adaptation, whereas the task objective, optimizer, evaluator, and execution interface remain fixed. The purpose of this setting is to study the interaction between two adaptive components of the learning system rather than unrestricted self-modification.

\subsection{Policy-State Characterization}
\label{sec:state}

A context should not be revised solely because another prompt obtains a higher reward under the current checkpoint. Such a criterion treats context optimization as an independent outer-loop search and does not explain how the current prompt has become mismatched with the evolving policy. We instead characterize the policy--context relationship along two complementary dimensions: behavioral exploration and instruction utilization. The first describes how broadly the policy distributes probability over possible continuations, while the second describes how strongly the policy uses different parts of the strategy scaffold.

We use policy entropy as the exploration signal. For assistant-generated token positions, we define
\begin{equation}
H_t(p)
=
\mathrm{E}*{x,\tau,j}
\left[
-\sum_v
\pi*{\theta_t}(v\mid x,p,y_{<j})
\log \pi_{\theta_t}(v\mid x,p,y_{<j})
\right].
\label{eq:entropy}
\end{equation}
A relatively low value of $H_t(p)$ indicates that the current policy is concentrated on a narrower set of continuations, whereas a relatively high value indicates broader behavioral exploration. We do not assume that a single absolute entropy value is optimal throughout training. Because the policy itself changes over time, the entropy of the current prompt is interpreted relative to historical mean. This produces a policy-relative exploration state and avoids treating global changes in model confidence as context-specific effects.

To measure instruction utilization, we partition the editable scaffold into semantic sections by splitting it into sentences. This avoids the poor generalization caused by dividing the scaffold into fixed sections.
We then estimate how much attention the generated response assigns to each section.
Let $I_k$ denote the prompt tokens belonging to section $k$, and let $\alpha_{j,i}$ denote the attention from response token $j$ to prompt token $i$. The utilization of section $k$ is summarized as
\begin{equation}
U_{t,k}(p)
=
\mathrm{E}*{x,\tau,j}
\left[
\frac{1}{|I_k|}
\sum*{i\in I_k}
\alpha_{j,i}
\right].
\label{eq:utilization}
\end{equation}
The length normalization prevents longer sections from receiving a systematically larger score simply because they contain more tokens. As with entropy, utilization is interpreted relative to measurements obtained from the current checkpoint rather than through a fixed threshold. In practice, attention can also be restricted to response stages associated with the semantic role of a section.

Entropy and utilization capture different aspects of the evolving learning state. Policy entropy indicates whether the current behavior is comparatively narrow or diffuse, while section-level utilization indicates whether the strategy guidance is being weakly or strongly used. Their combination therefore provides a policy-dependent signal for deciding both the direction and the form of a context intervention. We use attention as an operational proxy for instruction utilization, not as evidence that attention itself is causally responsible for model behavior~\citep{jain2019attention}.

\subsection{Policy-Aware Context Evolution}
\label{sec:context}
\begin{wrapfigure}{r}{0.48\linewidth}
    \centering
    \vspace{-0.8em}
    \includegraphics[width=\linewidth]{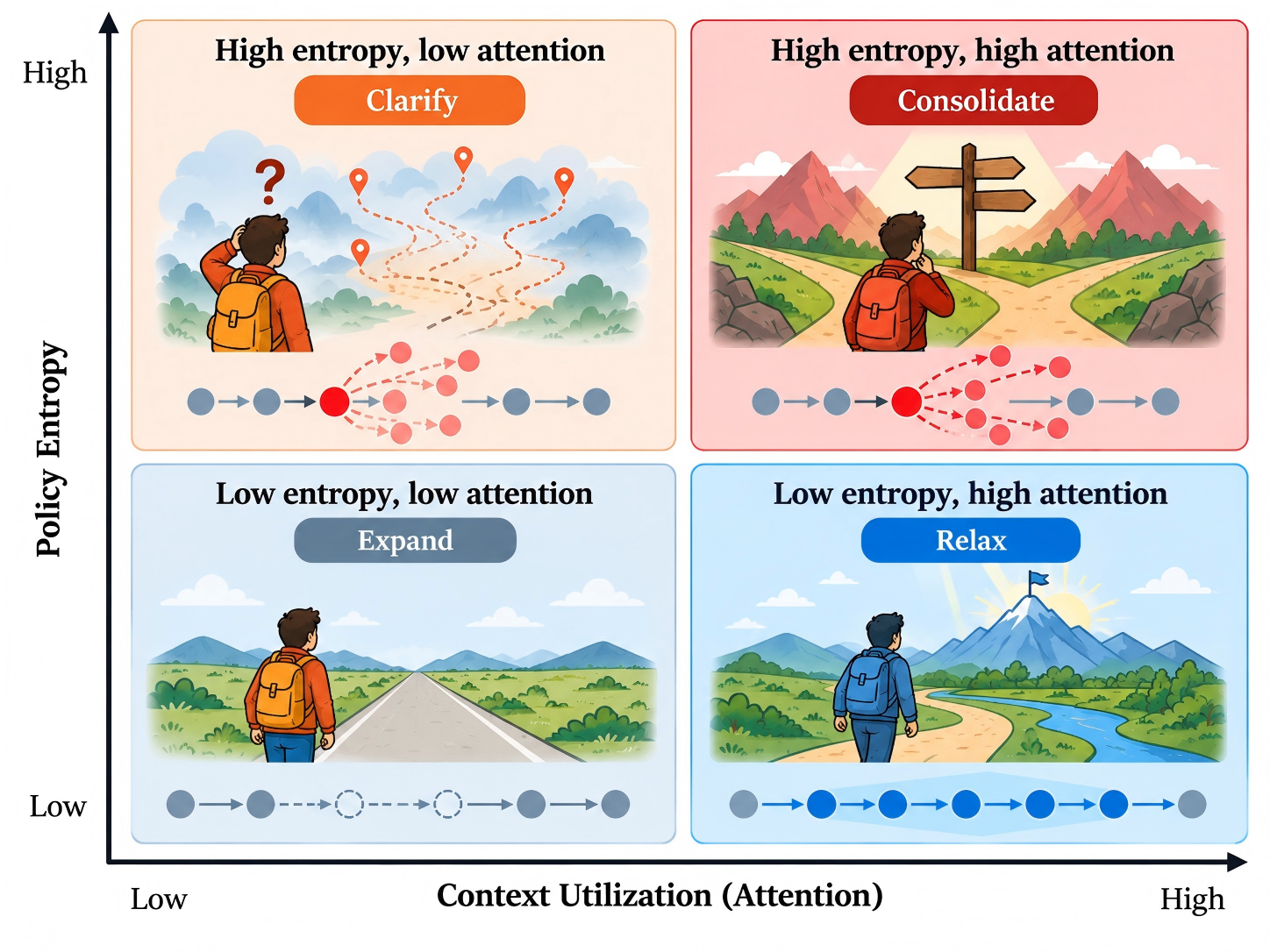}
    \caption{
        \textbf{Policy-aware context intervention.}
        Entropy measures behavioral exploration, while attention
        reflects policy-relative context utilization. Their combination
        determines how the strategy context is adapted to the current policy.
    }
    \label{fig:method}
    \vspace{-0.8em}
\end{wrapfigure}
At each context-update boundary, we characterize the current policy along two complementary dimensions: \emph{exploration}, measured by policy entropy, and \emph{utilization}, estimated from the policy's attention to the relevant strategy section. Exploration captures the diversity of the policy's current behavior, whereas utilization reflects how strongly that behavior is conditioned on the editable contextual guidance. Their joint state therefore provides a more informative signal for context adaptation than either quantity alone.

Figure~\ref{fig:method} shows the resulting entropy--attention state space. Intuitively, low exploration indicates that the policy has become behaviorally concentrated, while high exploration indicates that multiple behavioral modes remain active. Low utilization suggests that the relevant contextual guidance is exerting little influence on the policy, whereas high utilization indicates that the policy is strongly conditioned on that guidance. These two axes induce four qualitatively different intervention regimes, summarized in Table ~\ref{tab:intervention}.

\begin{table}[t]
\begingroup
\setlength{\tabcolsep}{3pt}
\renewcommand{\arraystretch}{1.15}
\begin{tabular}{@{}p{0.12\linewidth}p{0.11\linewidth}p{0.13\linewidth}p{0.58\linewidth}@{}}
\hline
\textbf{Exploration} &
\textbf{Utilization} &
\textbf{Intervention} &
\textbf{Context update} \\
\hline

Low &
Low &
\textbf{Expand} &
The policy is behaviorally concentrated while making limited use of the existing strategy guidance. Introduce an alternative strategy or scaffold to enlarge the available behavioral space. \\

Low &
High &
\textbf{Relax} &
The policy is concentrated while strongly relying on the current guidance. Soften overly prescriptive instructions allowing
alternative reasoning paths when appropriate. \\

High &
Low &
\textbf{Clarify} &
The policy explores broadly but makes weak use of the relevant guidance. Rewrite the selected section so that its intended strategy, priority, or activation condition is more explicit. \\

High &
High &
\textbf{Consolidate} &
The policy remains highly exploratory despite strong contextual utilization. Merge redundant or competing guidance into a more coherent strategy scaffold to reduce fragmentation. \\

\hline
\end{tabular}
\caption{
Policy-aware context intervention. Exploration is measured by policy entropy,
while utilization is estimated from attention to the relevant strategy section.
Both signals are interpreted relative to the current policy's calibrated
reference range.
}
\label{tab:intervention}

\endgroup
\vspace{-12pt}
\end{table}

The four regimes specify the \emph{direction} of context adaptation rather than an unconditional editing rule. We intervene only when the measured policy state provides sufficient evidence that a context update is warranted. In particular, if exploration remains within its calibrated reference range, or if the utilization estimate is insufficiently reliable, the incumbent prompt is preserved. When multiple editable sections exhibit substantial utilization deviations, the controller selects the section with the strongest policy-relative mismatch. The detailed editing prompts are provided in Appendix~\ref{app:prompts}.

Given a selected intervention, the current policy checkpoint is used as a constrained context editor. The editor receives the incumbent prompt, the target strategy section, and the intervention type, and proposes a small set of local revisions. Only the editable strategy scaffold may be modified; the task specification, output requirements, and other immutable instructions remain fixed. This restriction preserves the underlying objective and isolates context evolution from changes to the task itself.

Importantly, a revision is not accepted merely because its textual form appears consistent with the intended intervention. Context changes are policy dependent: adding an alternative strategy does not necessarily increase exploration, relaxing an instruction does not necessarily diversify behavior, and clarifying a strategy does not guarantee greater utilization. We therefore evaluate every candidate revision using the same frozen checkpoint from which the policy-state measurements were obtained. A candidate is accepted only if it shifts the relevant policy state in the intended direction while preserving task performance within a predefined tolerance. If no candidate satisfies these conditions, the incumbent prompt is retained. The complete training and context-evolution procedure is provided in Algorithm~\ref{alg:coevo-latest-step}(Appendix~\ref{app:coevo-algorithm}).

After a context revision is accepted, on-policy training resumes under the updated prompt. The resulting co-evolution cycle is

\begin{equation}
(\theta_t, p_t)
\longrightarrow
(H_t, U_t)
\longrightarrow
p_{t+1}
\longrightarrow
\mathcal{D}_{t+1}^{\mathrm{on}}
\longrightarrow
\theta_{t+1},
\label{eq:loop}
\end{equation}

where $(H_t,U_t)$ summarizes the interaction between the current policy and
its contextual strategy scaffold, $p_{t+1}$ denotes the verified context
revision, and $\mathcal{D}_{t+1}^{\mathrm{on}}$ is the subsequent on-policy
experience generated under that context. The context update thus changes the
experience distribution used for parameter learning, while the resulting
parameter update changes the policy state against which the next context
revision is evaluated. Repeating this process closes the feedback loop between
parameter learning and context evolution.
\section{Experiments}
\label{sec:experiments}
\paragraph{Models and domains.}
We evaluate Qwen3.5-4B, Qwen3.5-4B-Base, Qwen3.5-9B and Qwen3.8-27B on mathematical reasoning and code generation. Math models are trained on DAPO-Math-17k and evaluated on AIME 2025~\citep{yu2025dapo}. Code experiments use a temporal split derived from LiveCodeBench~\citep{jain2024livecodebench}. For each model scale, code RL starts from a common SFT checkpoint trained on 300 verified trajectories. Parameter updates use LoRA-based, GRPO-style optimization. Detailed training configurations and execution protocols are provided in Appendix~\ref{app:experiments}.

\paragraph{Baselines.}
We compare \method{} with fixed-prompt GRPO using Simple, Medium, and Detail training contexts. These baselines keep the context unchanged throughout RL, allowing us to assess the benefit of adapting contextual guidance during training. We further evaluate the resulting checkpoints under shared inference prompts to distinguish training effects from differences in evaluation-time context. Given the computational cost of full RL training, we restrict comparisons with E-SPL-style baselines to the ablation setting~\citet{zhang2026espl}.

\paragraph{Method Settings.} COEVO performs one context-evolution attempt every five RL steps. At each attempt, the current policy is frozen to guide context revision and evaluate candidate prompts. If no candidate satisfies the acceptance criteria, the current prompt is retained.

\paragraph{Evaluation.}
We report Average@12 and Pass@12 in both domains. Average@12 measures the mean per-problem success rate across samples, while Pass@12 measures the fraction of problems solved by at least one sample. Cross-prompt evaluation applies the same fixed evaluation prompts to each checkpoint, assessing both performance under shared contexts and sensitivity to prompt changes. Code results are macro-averaged over the three evaluation prompts and refer to our local temporal split rather than the official LiveCodeBench benchmark.

\section{Results and Analysis}
\label{sec:results}

\subsection{Main Results}
\label{sec:main-results}

\begin{figure}[t]
    \centering
    \includegraphics[width=\linewidth]{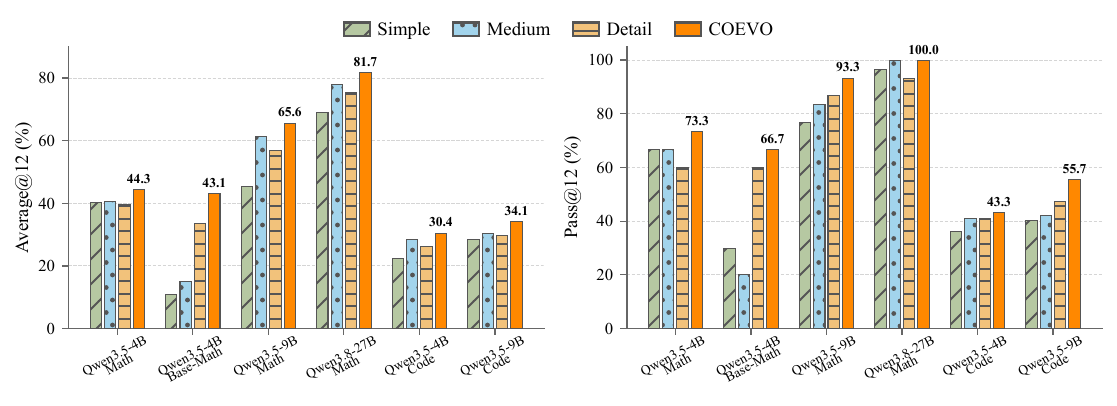}
    \caption{
    Performance comparison of COEVO with different prompt complexity levels across Qwen3.5 model scales and task domains. Simple, Medium, and Detail denote fixed prompts with increasing levels of prompt specification, while COEVO represents our method. 
    }
    \label{fig:result}
    \vspace{-15pt}
\end{figure}

\paragraph{Performance improvement.}
Figure~\ref{fig:result} compares \method{} with fixed-prompt RL across model scales and task domains. On Math, \method{} consistently improves both Average@12 and Pass@12 at 4B and 9B scales. The improvement in Average@12 indicates that co-evolution increases the overall quality of sampled responses rather than benefiting only a small subset of generations, while the simultaneous improvement in Pass@12 shows that the learned policy is also more likely to produce at least one successful solution within a fixed sampling budget. Importantly, stronger fixed prompts alone do not lead to monotonic improvements: increasing prompt detail can improve one metric while degrading another. In contrast, \method{} adapts the training context as the policy evolves and achieves the strongest overall performance, suggesting that the benefit comes from matching contextual guidance to the evolving policy rather than simply supplying more detailed instructions.

\paragraph{Robustness to inference prompts.}
Table~\ref{tab:crossprompt-4b} evaluates each trained policy under all three inference prompts. \method{} achieves the strongest worst-case performance under both metrics. Its worst-case Average@12 is 43.06\%, exceeding the strongest fixed-context baseline by 2.78 percentage points, while its worst-case Pass@12 reaches
73.33\%, a 6.66-point improvement. Moreover, Pass@12 is identical across all three inference prompts, indicating that task-level coverage
is insensitive to the evaluation prompt in this setting. For Average@12, the Detail-trained baseline exhibits a smaller max--min gap (0.83 versus 2.22), but at a substantially lower performance level (38.89 worst-case versus 43.06 for \method{}). Thus, low variation alone does not imply strong robust performance: \method{} combines high average performance with substantially stronger worst-case behavior across inference contexts. Overall, these results suggest that context evolution during training also
improves the learned policy's robustness to inference-time prompt variation.

\begin{table*}[t]
\centering
\small
\setlength{\tabcolsep}{7pt}
\renewcommand{\arraystretch}{1.05}

\begin{tabular}{@{}llrrrrrr@{}}
\toprule
\textbf{Metric} &
\textbf{Training} &
\textbf{Simple} &
\textbf{Medium} &
\textbf{Detail} &
\textbf{Mean $\uparrow$} &
\textbf{Worst $\uparrow$} &
\textbf{$\Delta$ $\downarrow$} \\
\midrule

Average@12
& Simple
& 44.72 & 40.28 & 40.28
& 41.76 & 40.28 & 4.44 \\

& Medium
& 42.78 & 41.94 & 40.28
& 41.67 & 40.28 & 2.50 \\

& Detail
& 39.17 & 38.89 & 39.72
& 39.26 & 38.89 & \textbf{0.83} \\

& \method{}
& 43.06 & 45.28 & 44.17
& \textbf{44.17}
& \textbf{43.06}
& 2.22 \\

\midrule

Pass@12
& Simple
& 66.67 & 70.00 & 66.67
& 67.78 & 66.67 & 3.33 \\

& Medium
& 63.33 & 70.00 & 66.67
& 66.67 & 63.33 & 6.67 \\

& Detail
& 63.33 & 63.33 & 60.00
& 62.22 & 60.00 & 3.33 \\

& \method{}
& 73.33 & 73.33 & 73.33
& \textbf{73.33}
& \textbf{73.33}
& \textbf{0.00} \\

\bottomrule
\end{tabular}
\caption{
\textbf{Cross-prompt robustness on 4B-Math.}
Models trained with different contexts are evaluated under the same
Simple, Medium, and Detail inference prompts.
\textbf{Worst} denotes the minimum performance across inference prompts,
and $\Delta$ denotes the max--min gap; higher Worst and lower $\Delta$
indicate stronger robustness to inference-prompt variation.
}
\label{tab:crossprompt-4b}

\end{table*}

\subsection{Case Study}
\label{sec:case-study}

To understand how the external context evolves with the policy, we analyze two representative training trajectories. Case~I shows how different policy states lead to different types of context revision, while Case~II provides complementary evidence that evolution can remove redundant guidance rather than continually adding instructions.

\paragraph{Context revision depends on the policy state.}
Case~I contains 20 attempted revisions, of which 11 are accepted. Table~\ref{tab:case-transitions} highlights representative accepted transitions. When exploration and section utilization are both low, the controller introduces additional structure: generic tool-use advice is replaced with explicit problem classification, and later expanded into symbolic, numerical, and search categories with a concrete tool-selection rule. When utilization becomes high under low exploration, the direction reverses and rigid requirements are relaxed. High-exploration states instead lead to stronger organization or consolidation of the existing strategy.

\begin{table}[t]
\centering
\footnotesize
\setlength{\tabcolsep}{4pt}
\renewcommand{\arraystretch}{1.02}
\begin{tabular}{@{}clp{0.67\linewidth}@{}}
\toprule
\textbf{Step} & \textbf{State (Ent./Util.)} & \textbf{Context revision} \\
\midrule
4  & L / L-L-L & Generic tool advice $\rightarrow$ explicit strategy selection \\
29 & L / L-L-L & Add problem categories and a cost--benefit decision rule \\
49 & H / L-L-L & Strengthen guidance into an explicit decision procedure \\
54 & L / H-H-H & ``must invoke'' $\rightarrow$ ``consider invoking'' \\
69 & L / L-L-H & Reintroduce classification and clearer tool-use conditions \\
99 & H / H-H-L & Consolidate guidance into a concise three-step procedure \\
\bottomrule
\end{tabular}
\caption{Representative accepted context revisions in Case~I.}
\label{tab:case-transitions}
\vspace{-12pt}
\end{table}

\paragraph{Context evolution is non-monotonic.}
COEVO does not simply make the prompt more detailed or restrictive. Early revisions add structure rules, while later revisions soften the same constraints once the policy relies heavily on them. Subsequent states can  trigger more explicit guidance before the strategy is consolidated. Moreover, only 11 of 20 proposed revisions are accepted, showing that context evolution is selective rather than the cumulative addition of editor-generated changes.

\paragraph{Evolution can also remove redundant guidance.}
Case~II exhibits a complementary pattern. The initial context contains detailed operational instructions about tool invocation and output formatting. During evolution, several of these details are removed or rewritten into more direct instructions while the core behavioral requirements remain intact. Table~\ref{tab:case-comparison} summarizes the two patterns.

\begin{table}[t]

\centering
\small
\begin{tabular}{@{}lp{0.35\linewidth}p{0.44\linewidth}@{}}
\toprule
\textbf{Case} & \textbf{Observed evolution} & \textbf{Implication} \\
\midrule
Case I & Structure $\rightarrow$ stronger constraints $\rightarrow$ relaxation $\rightarrow$ consolidation & Appropriate guidance changes with the current policy state. \\
Case II & Detailed operational guidance $\rightarrow$ more direct strategy instructions & Redundant guidance can be removed while preserving core task requirements. \\
\bottomrule
\end{tabular}

\caption{Complementary context-evolution patterns observed in the two case studies.}
\label{tab:case-comparison}
\vspace{-15pt}

\end{table}

Together, the two cases provide a concrete view of parameter--context co-evolution. Parameter updates change how contextual guidance is used, while context updates reshape the conditions under which subsequent experience is generated. The context therefore evolves toward different forms at different stages rather than toward a single fixed prompt structure.

\subsection{Ablation and Mechanism Analysis}
\label{sec:ablation}

We compare \method{} against two single-signal variants and an E-SPL-style
reward-driven baseline inspired by~\citet{zhang2026espl}.

\begin{wraptable}{r}{0.40\columnwidth}
\vspace{-3mm}
\centering
\small
\setlength{\tabcolsep}{4.5pt}
\renewcommand{\arraystretch}{1.0}
\begin{tabular}{@{}lccc@{}}
\toprule
\textbf{Method} &
\textbf{Avg.} &
\textbf{Pass} &
$\overline{H}$ \\
\midrule
Attn.-only
& 40.2 & 66.7 & 0.35 \\
Ent.-only
& 42.0 & 70.0 & 0.47 \\
E-SPL
& 41.6 & 66.7 & 0.31 \\
\method{}
& \textbf{44.3}
& \textbf{73.3}
& \textbf{0.52} \\
\bottomrule
\end{tabular}
\vspace{-1mm}
\caption{\footnotesize Ablation on Qwen3.5-4B Math.}
\label{tab:ablation}
\vspace{-3mm}
\end{wraptable}

Attention-only retains context-utilization feedback but removes the explicit
exploration signal. It can identify which parts of the scaffold are being
used, but cannot detect whether the policy is becoming overly concentrated;
correspondingly, its entropy decreases substantially over training.
Entropy-only exhibits the complementary failure mode: it observes changes in
exploration but cannot localize how the current scaffold is being utilized,
leading to weaker downstream performance. Together, these results suggest
that entropy and attention provide complementary information for selecting
context interventions.

The E-SPL-style baseline reveals a different failure mode. Although it
achieves stronger on-policy performance during training, its held-out
Average@12 and Pass@12 reach only 41.61\% and 66.67\%, trailing \method{}
by 2.70 and 6.66 points, respectively. This performance gap coincides with
a larger reduction in policy entropy and output diversity, consistent with
premature exploitation: reward-driven context updates can reinforce
immediately successful strategies while narrowing the policy before learning
is complete. In contrast, \method{} conditions context revision on both
exploration and utilization, allowing the scaffold to adapt while preserving
behavioral diversity.

\section{Related Work}
\paragraph{Reinforcement learning and policy-state characterization.}
Reinforcement learning has become an important mechanism for improving reasoning and decision-making in language models. GRPO and DAPO develop scalable on-policy optimization methods, while ReTool studies reinforcement learning with execution feedback~\citep{shao2024deepseekmath,yu2025dapo,feng2025retool}. Recent work further explores adaptive training curricula and self-play for continued policy improvement~\citep{ye2025eva}. Studies of policy entropy reveal its relationship with exploration, reasoning behavior, and premature convergence during RL~\citep{cui2025entropy,cheng2026reasoning,hao2026rethinking,xu2026entropyflow}. Meanwhile, attention provides a localized, though interpretation-sensitive, view of how models utilize instructional context~\citep{jain2019attention,wiegreffe2019attention}. We use policy entropy and prompt-conditioned attention as complementary descriptors of the evolving policy state, capturing behavioral exploration and instruction utilization, respectively. Rather than optimizing these signals as standalone objectives, we use them to guide context evolution during reinforcement learning.

\paragraph{Recursive self-improvement.}
Recent work on recursive self-improvement explores how language models can iteratively improve their external context, learning experience, or computational structures. Reflexion and CURE study feedback-driven refinement and iterative self-improvement~\citep{shinn2023reflexion,chen2026cure}, while OPRO, EvoPrompt, Promptbreeder, GEPA, and TextGrad optimize prompts or system components through search, evolutionary operators, or textual feedback~\citep{yang2023opro,guo2023evoprompt,fernando2023promptbreeder,agrawal2025gepa}. LSE trains models to evolve their contexts at test time, while in-context RL demonstrates improvement from reward-conditioned interaction histories~\citep{chen2026lse,song2026reward}. SPEE combines progressive experience evolution with policy optimization~\citep{ren2026spee}. Beyond prompt-level adaptation, STOP, Self-Developing, and the Darwin Godel Machine investigate self-improvement through program, algorithm, or agent-code modification~\citep{zelikman2023stop,ishibashi2025selfdeveloping,zhang2025dgm}. E-SPL couples system-prompt evolution with reinforcement-learning-based parameter updates~\citep{zhang2026espl}. Our work focuses on policy-aware parameter--context co-evolution: entropy and attention guide system-context adaptation during RL, while the surrounding optimization and evaluation machinery remains fixed.

\section{Conclusion}
This work presents parameter--context co-evolution as a broader perspective for understanding recursive self-improvement in LLMs. Rather than viewing RSI solely as repeated optimization in parameter space, we argue that the external context shaping model behavior and learning can itself become part of the adaptive process. This perspective highlights RSI as the coordinated evolution of multiple interacting components, where improvements in the model change how contextual guidance is used, while changes in context reshape the experience that drives further learning. By studying this interaction within a bounded and controllable setting, our framework provides a useful foundation for analyzing more general forms of self-improving systems and for understanding how different adaptive mechanisms may jointly contribute to recursive capability development.

\label{page:main-end}
\clearpage

\bibliography{references}
\bibliographystyle{plainnat}

\clearpage
\appendix
\providecommand{\method}{\textsc{CoEvo}}
\providecommand{\tbd}[1]{\textcolor{red}{\textbf{[#1]}}}
\providecommand{\pendingtext}[1]{{\color{red}#1}}

\section{Experimental Setup}
\label{app:experiments}

\subsection{Models and tasks}

\begin{table}[h]
\centering
\small
\begin{tabularx}{\linewidth}{@{}llYY@{}}
\toprule
\textbf{Model} & \textbf{Domain} & \textbf{Training data} & \textbf{Evaluation data} \\
\midrule
Qwen 3.5-4B & Math & DAPO-Math-17k & AIME 2025 \\
Qwen 3.5-4B-Base & Math & DAPO-Math-17k & AIME 2025 \\
Qwen 3.5-9B & Math & DAPO-Math-17k & AIME 2025 \\
Qwen 3.8-27B & Math & DAPO-Math-17k & AIME 2025 \\
Qwen 3.5-4B & Code & LCB-v6 temporal train & 100-task stratified subset of frozen 200-task test split \\
Qwen 3.5-9B & Code & LCB-v6 temporal train & 100-task stratified subset of frozen 200-task test split \\
\bottomrule
\end{tabularx}
\caption{Experimental settings.}
\label{tab:app-settings}
\end{table}

For Math, after holding out a 50-problem development split, the training set contains
17{,}867 unique problems, and AIME 2025 contains 30 evaluation problems. We use the tokenizer and chat template
bundled with each checkpoint.

The Code experiments use a fixed temporal split derived from the LiveCodeBench
release-v6 data. After excluding 175 pilot-task IDs, the train, calibration,
checkpoint-development, and final-test splits contain 400, 60, 100, and 200
mutually disjoint tasks, ordered by release time. The same 100 task IDs are used
for both model scales. They are sampled without replacement from the frozen
200-task test split using proportional difficulty stratification, yielding
23 easy, 33 medium, and 44 hard tasks. Both model scales use a shared SFT
checkpoint trained for 3 epochs on 300 verified trajectories.

\subsection{RL configuration}

\begin{table}[h]
\centering
\small
\begin{tabularx}{\linewidth}{@{}lYY@{}}
\toprule
\textbf{Setting} & \textbf{Math} & \textbf{Code} \\
\midrule
Initialization
& Qwen 3.5-4B, Qwen 3.5-4B-Base,
  Qwen 3.5-9B, and Qwen 3.8-27B
& Qwen 3.5-4B and Qwen 3.5-9B
  initialized from a shared SFT checkpoint \\
RL algorithm
& GRPO-style group-mean advantage without standard-deviation normalization;
  PPO clip $[0.8,1.28]$
& GRPO-style group-mean advantage without standard-deviation normalization;
  PPO clip $[0.8,1.28]$ \\
Optimizer
& Adam, $\beta_1=0.9$, $\beta_2=0.95$
& Adam, $\beta_1=0.9$, $\beta_2=0.95$ \\
Learning rate
& $4\times10^{-5}$
& $4\times10^{-5}$ \\
Batch size (task per step) 
& 8 (4B), 4 (9B, 27B)
& 4  \\
Rollouts per task 
& 8
& 8 \\
Training steps
& 100
& 100 \\
Training sampling
& Temperature 1.0, top-$p$ 1.0
& Temperature 1.0, top-$p$ 1.0 \\
Checkpoint interval
& 50 steps (4B), 25 steps (9B, 27B)
& 20 steps \\
Training seed
& 42
& 42 \\
LoRA rank
& 32
& 32 \\
\bottomrule
\end{tabularx}
\caption{RL training hyperparameters.}
\label{tab:app-rl}
\end{table}

Math trajectories allow at most 6 assistant turns, at most 4
(4B) or 3 (9B, 27B) code-interpreter calls, 512 tokens per tool response,
and a 30-second sandbox timeout. The maximum trajectory length is 8192
tokens for 4B and 32768 tokens for 9B and 27B, with per-turn limits of
1024 and 4096 tokens, respectively.

Code trajectories allow at most 2 assistant turns, 1 isolated and stateless
\texttt{run\_python} call, 10{,}240 tokens per assistant response, 20{,}480
total trajectory tokens, and 512 tool-response tokens. 

\subsection{Evaluation protocol}

Math and Code are evaluated using Average@12 and Pass@12. For each task,
twelve responses are sampled under the same evaluation configuration.

Let $x$ denote an evaluation task and $z_{x,r}\in\{0,1\}$ indicate whether
the $r$-th sampled response for task $x$ is successful. Then

\begin{equation}
\operatorname{Average@12}(x)
=
\frac{1}{12}
\sum_{r=1}^{12}
z_{x,r},
\label{eq:app-average12}
\end{equation}

\begin{equation}
\operatorname{Pass@12}(x)
=
\mathbf{1}
\left[
\sum_{r=1}^{12}z_{x,r}>0
\right].
\label{eq:app-pass12}
\end{equation}

For Math, the 4B evaluation uses temperature 1.0 and top-$p$ 0.7, while the 9B
and 27B evaluations use temperature 0.7 and top-$p$ 0.9. The corresponding
trajectory limits are 8192 tokens for 4B and 32768 tokens for 9B and 27B, with
per-turn limits of 1024 and 4096 tokens, respectively. Generation stops when a
\verb|\boxed{}| answer is emitted or when the trajectory budget is exhausted.
For answer checking, the last \verb|\boxed{...}| group, including nested braces,
is extracted; whitespace, commas, and \texttt{\$} are removed before exact integer
comparison with the AIME ground truth.

For Code, evaluation uses temperature 1.0, top-$p$ 0.7, at most 10{,}240 tokens
per assistant response, two assistant turns, one isolated and stateless
\texttt{run\_python} call, 20{,}480 total trajectory tokens, and 512 tool-response
tokens. The tool-call timeout is 5 seconds, the per-test-case timeout
is 3 seconds, and the whole-problem judge timeout is 30 seconds.

For Code, a response is successful only when the submitted program passes the full
judge. Average@12 is the empirical success rate across the twelve generations, while Pass@12 is the
fraction of tasks for which at least one generation succeeds.

\subsection{Context-Evolution Algorithm}
\label{app:coevo-algorithm}

\paragraph{Update schedule and data reuse.}
Let $\theta_t$ and $p_t=c\oplus s_t$ be the policy and prompt before
zero-based RL step $t$. Here $c$ contains the fixed task specification,
tool interface, and output constraints, and $s_t$ is the editable strategy
scaffold. At step $t$, the model samples training trajectories for task batch
$\mathcal B_t$ under $p_t$ and performs its ordinary RL update.
Context evolution is attempted whenever $(t+1)\bmod 5=0$, i.e.,
after zero-based steps $4,9,\ldots$.

The controller uses the statistics and task examples from the last training
step, rather than pooling the preceding five steps. The updated checkpoint
$\bar\theta=\theta_{t+1}$ is frozen during prompt generation and comparison.
Candidate prompts are tested on the same latest task examples
$\mathcal B_t$ under this checkpoint. The five-step interval determines
when editing is attempted; it does not define an additional evaluation set.

\paragraph{Historical reference used in the implementation.}
Let $h_t$ and $u_{t,k}$ denote the entropy and utilization
of strategy section $k$ recorded in the latest training step.
The controller compares these values with their cumulative
historical means, including the current step:
\begin{equation}
\mu^H_t = \frac{1}{t+1}\sum_{j=0}^{t}h_j,
\qquad
\mu^U_{t,k} = \frac{1}{t+1}\sum_{j=0}^{t}u_{j,k}.
\label{eq:coevo-history}
\end{equation}
These historical means are computed from training records
collected as the policy evolves. They are not obtained by
evaluating a fixed set of reference prompts under one checkpoint.
At the first context-update boundary, both exploration and
utilization states are initialized to Low.

The frozen-checkpoint comparison is a separate operation:
after generating candidate prompts, the incumbent and candidates
are evaluated under the current frozen checkpoint on the same
latest training batch. This comparison is used for candidate
selection, not for constructing the historical reference.

\paragraph{Generation and comparison.}
The current frozen checkpoint proposes eight edited strategy scaffolds.
Each candidate preserves the immutable component $c$.
Let $\mathcal C_t$ contain the incumbent $p_t$ and the valid edited prompts.
For each $p\in\mathcal C_t$, generate fresh responses to the same last-step
task examples under $\pi_{\bar\theta}(\cdot\mid x,p)$ and compute the
comparison statistics:
\begin{equation}
 z_t(p)=
 \bigl(\widehat H_t(p),\widehat{\mathbf U}_t(p),\widehat R_t(p)\bigr).
 \label{eq:coevo-candidate-measure}
\end{equation}
The incumbent is re-evaluated under the same checkpoint rather than being
compared using its old training trajectories. Candidate comparisons reuse
the existing rollout count and decoding configuration; there is no separate
verification-batch-size or verification-rollout-count hyperparameter.
The evaluation task IDs and actual generation settings must be recorded.

The next prompt is the candidate ranked best in the requested direction:
\begin{equation}
 p_{t+1}
 =\operatorname{BestInDirection}
   \bigl(\mathcal C_t,\{z_t(p):p\in\mathcal C_t\},d_t\bigr).
 \label{eq:coevo-select}
\end{equation}

\begin{algorithm}[t]
\caption{COEVO with latest-step prompt comparison}
\label{alg:coevo-latest-step}
\begin{algorithmic}[1]
\Require Initial checkpoint $\theta_0$, prompt $p_0=c\oplus s_0$,
 training data, RL configuration, number of steps $T$
\State Initialize the training-statistics history
\For{$t=0,\ldots,T-1$}
  \State Sample task batch $\mathcal B_t$
  \State $\mathcal T_t\gets\operatorname{Rollout}(\theta_t,p_t,\mathcal B_t)$
  \State Record latest-step entropy $h_t$ and section utilization $\mathbf u_t$
  \State Append these statistics to the historical records
  \State $\theta_{t+1}\gets\operatorname{RLUpdate}(\theta_t,\mathcal T_t)$
  \State $p_{t+1}\gets p_t$
  \If{$(t+1)\bmod 5=0$}
    \State Compute historical means, including the current step
    \State Compare $(h_t,\mathbf u_t)$ with the historical means
    \State At the first boundary, initialize both states to Low
    \State Determine edit direction $d_t$ and section instructions
    \State Freeze $\bar\theta\gets\theta_{t+1}$
    \State Generate eight candidate edits using $\bar\theta$ and Appendix~\ref{app:prompts}
    \State $\mathcal C_t\gets\{p_t\}\cup\{\text{valid edited prompts}\}$
    \For{each $p\in\mathcal C_t$}
      \State Sample responses under $(\bar\theta,p)$ on the same latest tasks $\mathcal B_t$
      \State Compute $z_t(p)$ using Eq.~\eqref{eq:coevo-candidate-measure}
    \EndFor
    \State $p_{t+1}\gets\operatorname{BestInDirection}(\mathcal C_t,\{z_t(p)\},d_t)$
  \EndIf
\EndFor
\State \Return $(\theta_T,p_T)$
\end{algorithmic}
\end{algorithm}

\paragraph{State carried to the next step.}
Only the selected strategy scaffold changes; the task interface and reward
definition remain fixed. The next RL step collects fresh on-policy trajectories
under the selected prompt. A context update performed after the final RL step
can change the returned prompt but cannot change the already trained weights.

\section{Prompts}
\label{app:prompts}
\definecolor{promptbg}{RGB}{248,249,251}
\lstset{
  basicstyle=\ttfamily\small,
  backgroundcolor=\color{promptbg},
  breaklines=true,
  columns=fullflexible,
  frame=single,
  framerule=0.35pt,
  framesep=6pt,
  keepspaces=true,
  rulecolor=\color{black!22},
  showstringspaces=false,
  aboveskip=6pt,
  belowskip=8pt
}
\subsection{Fixed prompts}

\subsubsection{Math prompts}

The following text is the protocol used in the reported math runs.

\paragraph{Math Simple: complete system prompt.}
\begin{lstlisting}
You solve math problems step by step. When you need to compute something,
write Python code in a code block like this:
```python
# your code here
print(result)
```
After you write a code block, I will execute it and show you the output.
Then you can continue reasoning. Put your final answer in \boxed{}.
\end{lstlisting}

\paragraph{Math Medium: complete system prompt.}
\begin{lstlisting}
You solve math problems step by step with help from a Python code interpreter.
Use the code_interpreter tool when calculation, symbolic manipulation, or
enumeration helps you solve the problem accurately and quickly.

How to use the code_interpreter tool:
- Call code_interpreter with a `code` string containing Python code.
  Call it at most once per assistant turn, then wait for the execution result.
- Results are captured from what your code prints with print().
  Always print the values you want to see.
- Each execution is independent: no variables, files, or state carry over
  between calls. Redefine everything you need in each piece of code.
- Code must finish within a few seconds and use little memory.
  Do not read or write files. If you enumerate or brute-force,
  keep the search space small.
- If the execution returns an error, analyze it and retry with corrected
  code when useful.

When you have the final answer, end with exactly one line in this format:
\boxed{<your final answer>}
Do not call the tool and give the final answer in the same turn.
\end{lstlisting}

\paragraph{Math Detail: complete system prompt.}
\begin{lstlisting}
You solve math problems step by step with help from a Python code interpreter.
Use the code_interpreter tool when calculation, symbolic manipulation, or
enumeration helps you solve the problem accurately and quickly.

To call the tool, output EXACTLY this format and then STOP immediately:
<tool_call>
<function=code_interpreter>
<parameter=code>
# python code, use print() to show results
print(result)
</parameter>
</function>
</tool_call>
Do not write anything after </tool_call>. The system will execute it and
show you the result.

How to use the code_interpreter tool:
- Call code_interpreter with a `code` string containing Python code.
  Call it at most once per assistant turn, then wait for the execution result.
- Results are captured from what your code prints with print().
  Always print the values you want to see.
- Each execution is independent: no variables, files, or state carry over
  between calls. Redefine everything you need in each piece of code.
- Code must finish within a few seconds and use little memory.
  Do not read or write files. If you enumerate or brute-force,
  keep the search space small.
- If the execution returns an error, analyze it and retry with corrected
  code when useful.

When you have the final answer, end with exactly one line in this format:
\boxed{<your final answer>}
Do not call the tool and give the final answer in the same turn.
\end{lstlisting}

\subsubsection{Code prompts}

The following text is the protocol used in the reported code runs.

\paragraph{Code Simple: complete system prompt.}
\begin{lstlisting}
You are an expert Python competitive-programming agent. Solve the user's problem with a correct and efficient Python 3.11 program using only the standard library. Follow the supplied starter code when present; otherwise read from stdin and write to stdout.

You may call run_python to execute code with optional stdin. Each call is isolated and stateless. Call it at most once per assistant turn and wait for the result.

When the solution is ready, return exactly one fenced Python code block and no other text.

You may call run_python at most 1 times in total. A tool call must be a short, focused check rather than a full solution or a reasoning scratchpad; omit long comments and exploratory code. When the tool budget is exhausted, or when further testing is unnecessary, immediately submit the final code in the required format.
\end{lstlisting}

\paragraph{Code Medium: text appended to “Code Simple”.}
\begin{lstlisting}
Reason concisely about the input/output, constraints, algorithm, and edge cases, then prioritize submitting the final implementation directly. Use run_python only for one short, focused check that resolves a specific uncertainty; never use it to draft the full solution or as a reasoning scratchpad. Reserve most of the response budget for the final program.
\end{lstlisting}

\paragraph{Code Detail: text appended to “Code Simple”.}
\begin{lstlisting}
Follow this workflow:

1. Identify the required input/output behavior, constraints, and important edge cases.
2. Derive a correct algorithm and verify its time and space complexity.
3. Implement the algorithm carefully in the required starter-code or stdin/stdout format.
4. Use run_python to check the provided examples when useful.
5. Check important boundary or adversarial cases, and fix the code if a check fails.
6. Review correctness, complexity, and output formatting before submitting the final code.
\end{lstlisting}

\paragraph{\texttt{run\_python} tool schema.}
\begin{lstlisting}
{
  "type": "function",
  "function": {
    "name": "run_python",
    "description": "Execute a short, focused Python 3.11 diagnostic in a fresh isolated container. Do not use the tool as a scratchpad or include long explanatory comments. Optional stdin is passed to the program. The call returns exit status, stdout, stderr, and timeout information. No files or state persist.",
    "parameters": {
      "type": "object",
      "properties": {
        "code": {"type": "string", "description": "Python code to execute."},
        "stdin": {"type": "string", "description": "Optional standard input."}
      },
      "required": ["code"]
    }
  }
}
\end{lstlisting}

The chat-template call uses \texttt{tools=[run\_python]}, \texttt{add\_generation\_prompt=True}, and \texttt{enable\_thinking=False}. The final answer is accepted only when it contains exactly one non-empty fenced Python code block.

\subsection{Prompt-evolution editor prompts}

The editor LLM receives a JSON request whose direction field selects one of four mutation strategies. The system message is always ``You edit system prompts. Return valid JSON only.''

\paragraph{Request skeleton.}
\begin{lstlisting}
{
  "task": "Generate alternative system prompts.",
  "direction": "<Clarify|Consolidate|Expand|Relax>",
  "strategy": "<see below>",
  "span_instructions": {
    "<section_key>": "TARGET|MINOR EDIT: <reason>. <instruction>",
    ...
  },
  "rules": [
    "Return exactly one JSON object with the same section keys.",
    "Each candidate must be a distinct rewrite.",
    "TARGET spans get the full edit; MINOR EDIT spans get minimal changes.",
    "Preserve the output format (e.g. boxed answer).",
    "Do not mention metrics, entropy, attention, or this request.",
    "No markdown fence -- return raw JSON only."
  ],
  "sections": { "<key>": "<current text>", ... }
}
\end{lstlisting}

\paragraph{Strategy texts.}

\begin{description}
  \item[Expand] Model ignores this section and is too deterministic. Convert abstract instructions into concrete strategy choices. Add missing thinking directions or decision frameworks.

  \item[Relax] Model attends to this section but is too deterministic. Change absolute requirements into conditional suggestions. Allow alternative paths. Reduce rigid binding.

  \item[Clarify] Model ignores this section and explores randomly. Rewrite vague instructions into clear goals with explicit priorities and decision conditions.

  \item[Consolidate] Model attends to this section but explores randomly. Merge duplicates, eliminate conflicts, clarify priority and conditions.
\end{description}

\paragraph{Span-instruction logic.}
For each section, the instruction is determined by the attention status and the direction:
\begin{lstlisting}
if direction in (scaffold, strengthen):       # low-attention operations
    if att == low:   "TARGET: ... <strategy>"
    if att == high:  "MINOR EDIT: ... keep mostly unchanged"
else:                                          # high-attention operations
    if att == high:  "TARGET: ... <strategy>"
    if att == low:   "MINOR EDIT: ... keep mostly unchanged"
\end{lstlisting}

\paragraph{Sampling parameters.}
temperature=0.8, top\_p=0.95, max\_tokens=1024, num\_samples = candidate count $-$ 1 (default 8).

\section{Additional Experimental Results}
\label{app:additional-results}

\subsection{Math Training Trajectories}

\subsubsection{Qwen3.5-4B-Base}

\begin{table}[t]
\centering
\small
\setlength{\tabcolsep}{7pt}
\renewcommand{\arraystretch}{1.05}
\begin{tabular}{@{}lcrrrr@{}}
\toprule
\textbf{Protocol} &
\textbf{Step} &
\textbf{Avg.@12} &
\textbf{Pass@12} &
\textbf{Format} &
\textbf{Code calls} \\
\midrule
Simple
& 0   &  2.50 & 16.67 &  2.78 & 0.03 \\
& 50  & 12.78 & 30.00 & 22.78 & 0.01 \\
& 100 & 10.83 & 20.00 & 33.89 & 0.10 \\
\addlinespace[3pt]
Medium
& 0   &  1.39 &  6.67 &  1.67 & 0.01 \\
& 50  & 11.11 & 23.33 & 17.50 & 0.00 \\
& 100 & 15.00 & 30.00 & 39.17 & 0.00 \\
\addlinespace[3pt]
Detail
& 0   &  6.67 & 30.00 & 27.22 & 2.34 \\
& 50  & 21.39 & 40.00 & 54.44 & 0.52 \\
& 100 & 33.61 & 60.00 & 49.17 & 1.70 \\
\bottomrule
\end{tabular}
\caption{Qwen3.5-4B-Base training trajectory on AIME 2025.}
\label{tab:app-4b-base}
\end{table}

Table~\ref{tab:app-4b-base} reports the Qwen3.5-4B-Base math training trajectory
through step 100 under the three tool-use protocols. Format and Code calls are
auxiliary diagnostics and are reported separately from the two task-level
performance metrics.

\subsubsection{Qwen3.5-9B}

\begin{table}[t]
\centering

\small
\setlength{\tabcolsep}{9pt}
\renewcommand{\arraystretch}{1.04}
\begin{tabular}{@{}lcrr@{}}
\toprule
\textbf{Protocol} &
\textbf{Step} &
\textbf{Avg.@12} &
\textbf{Pass@12} \\
\midrule
Simple
& 0 & 31.94 & 53.33 \\
& 25   & 43.61 & 73.33 \\
& 50   & 50.28 & 83.33 \\
& 75   & 51.67 & 90.00 \\
& 100  & 45.28 & 76.67 \\
\addlinespace[3pt]
Medium
& 0 & 49.44 & 76.67 \\
& 25   & 49.44 & 83.33 \\
& 50   & 61.39 & 86.67 \\
& 75   & 61.94 & 96.67 \\
& 100  & 61.39 & 83.33 \\
\addlinespace[3pt]
Detail
& 0 & 40.83 & 70.00 \\
& 25   & 53.89 & 86.67 \\
& 50   & 51.11 & 80.00 \\
& 75   & 57.78 & 86.67 \\
& 100  & 56.94 & 86.67 \\
\addlinespace[3pt]
\method{}
& 100 & \textbf{65.56} & 93.33 \\
\bottomrule
\end{tabular}
\caption{Qwen3.5-9B Math training trajectory on AIME 2025.}
\label{tab:app-math-scale-trajectories}
\end{table}

Table~\ref{tab:app-math-scale-trajectories} reports the Qwen3.5-9B math training trajectory. The reported \method{} aggregate is 65.56\%
Average@12 and 93.33\% Pass@12, achieving the highest.

\subsection{Code Checkpoint Selection}
\label{app:code-checkpoint-selection}

\begin{table}[!t]
\centering
\footnotesize
\setlength{\tabcolsep}{4.8pt}
\renewcommand{\arraystretch}{1.00}
\begin{tabular}{@{}lcrrrrrr@{}}
\toprule
\textbf{Train} &
\textbf{Step} &
\textbf{S} &
\textbf{M} &
\textbf{D} &
\textbf{Mean} &
\textbf{Worst} &
\textbf{Case} \\
\midrule
Simple & 20  & 53.00 & 53.00 & 59.00 & 55.00 & 53.00 & 62.41 \\
\textbf{Simple} & \textbf{40} &
\textbf{57.00} & \textbf{57.00} & \textbf{56.00} &
\textbf{56.67} & \textbf{56.00} & \textbf{64.05} \\
Simple & 60  & 48.00 & 52.00 & 55.00 & 51.67 & 48.00 & 61.46 \\
Simple & 80  & 60.00 & 55.00 & 55.00 & 56.67 & 55.00 & 65.95 \\
Simple & 100 & 56.00 & 58.00 & 54.00 & 56.00 & 54.00 & 63.78 \\
\addlinespace[2pt]
Medium & 20  & 48.00 & 49.00 & 48.00 & 48.33 & 48.00 & 57.61 \\
Medium & 40  & 57.00 & 59.00 & 53.00 & 56.33 & 53.00 & 65.19 \\
Medium & 60  & 63.00 & 61.00 & 60.00 & 61.33 & 60.00 & 67.75 \\
Medium & 80  & 59.00 & 61.00 & 60.00 & 60.00 & 59.00 & 66.50 \\
\textbf{Medium} & \textbf{100} &
\textbf{59.00} & \textbf{64.00} & \textbf{63.00} &
\textbf{62.00} & \textbf{59.00} & \textbf{68.40} \\
\addlinespace[2pt]
Detail & 20  & 53.00 & 51.00 & 55.00 & 53.00 & 51.00 & 60.89 \\
Detail & 40  & 47.00 & 50.00 & 52.00 & 49.67 & 47.00 & 56.61 \\
Detail & 60  & 52.00 & 57.00 & 57.00 & 55.33 & 52.00 & 61.99 \\
Detail & 80  & 53.00 & 59.00 & 56.00 & 56.00 & 53.00 & 62.95 \\
\textbf{Detail} & \textbf{100} &
\textbf{63.00} & \textbf{67.00} & \textbf{63.00} &
\textbf{64.33} & \textbf{63.00} & \textbf{71.76} \\
\bottomrule
\end{tabular}
\caption{Qwen3.5-4B Code Training Trajectory on LCB-v6.}
\label{tab:app-code-checkpoint-selection}
\label{tab:app-code-4b-checkpoint-selection}
\end{table}

\begin{table}[!t]
\centering
\footnotesize
\setlength{\tabcolsep}{4.8pt}
\renewcommand{\arraystretch}{1.00}
\begin{tabular}{@{}lcrrrrrr@{}}
\toprule
\textbf{Train} &
\textbf{Step} &
\textbf{S} &
\textbf{M} &
\textbf{D} &
\textbf{Mean} &
\textbf{Worst} &
\textbf{Case} \\
\midrule
Simple & 20  & 56.00 & 49.00 & 54.00 & 53.00 & 49.00 & 61.47 \\
Simple & 40  & 59.00 & 61.00 & 56.00 & 58.67 & 56.00 & 64.37 \\
Simple & 60  & 57.00 & 60.00 & 60.00 & 59.00 & 57.00 & 65.53 \\
Simple & 80  & 55.00 & 61.00 & 56.00 & 57.33 & 55.00 & 61.21 \\
\textbf{Simple} & \textbf{100} &
\textbf{60.00} & \textbf{60.00} & \textbf{58.00} &
\textbf{59.33} & \textbf{58.00} & \textbf{64.27} \\
\addlinespace[2pt]
Medium & 20  & 59.00 & 59.00 & 56.00 & 58.00 & 56.00 & 64.04 \\
Medium & 40  & 66.00 & 63.00 & 64.00 & 64.33 & 63.00 & 70.00 \\
\textbf{Medium} & \textbf{60} &
\textbf{71.00} & \textbf{69.00} & \textbf{69.00} &
\textbf{69.67} & \textbf{69.00} & \textbf{76.21} \\
Medium & 80  & 70.00 & 65.00 & 73.00 & 69.33 & 65.00 & 77.70 \\
Medium & 100 & 66.00 & 65.00 & 67.00 & 66.00 & 65.00 & 74.65 \\
\addlinespace[2pt]
Detail & 20  & 55.00 & 56.00 & 53.00 & 54.67 & 53.00 & 62.89 \\
Detail & 40  & 60.00 & 60.00 & 54.00 & 58.00 & 54.00 & 66.20 \\
\textbf{Detail} & \textbf{60} &
\textbf{64.00} & \textbf{65.00} & \textbf{71.00} &
\textbf{66.67} & \textbf{64.00} & \textbf{72.75} \\
Detail & 80  & 67.00 & 61.00 & 67.00 & 65.00 & 61.00 & 69.93 \\
Detail & 100 & 69.00 & 62.00 & 65.00 & 65.33 & 62.00 & 70.26 \\
\bottomrule
\end{tabular}
\caption{Qwen3.5-9B Code Training Trajectory on LCB-v6.}
\label{tab:app-code-9b-checkpoint-selection}
\end{table}

Table~\ref{tab:app-code-4b-checkpoint-selection} and
Table~\ref{tab:app-code-9b-checkpoint-selection} report the Qwen3.5-4B and
Qwen3.5-9B code training trajectories on LCB-v6, respectively. S/M/D denote the
All-Case Pass Rate under the Simple/Medium/Detail inference prompts, i.e., the
percentage of tasks for which all test cases pass. This differs from Case, which
is the mean test-case pass rate and gives partial credit, e.g., 3/5 test cases
passed yields 0.6. All entries other than Step are percentages. Checkpoints are selected
on a separate 100-task development split, where each candidate is evaluated
under all three inference prompts with one greedy response per task and prompt.
Selection maximizes Mean, then Worst, then Case; ties are broken by the earlier
step. This yields Simple step~40, Medium step~100, and Detail step~100 for
4B-Code, and Simple step~100, Medium step~60, and Detail step~60 for 9B-Code.
The unequal selected steps therefore reflect development-set model selection
rather than unequal training budgets.

\subsection{Final Code Evaluation}
\label{app:code-results}
\label{app:4b-code-results}
\label{app:9b-code-results}

\begin{table}[t]
\centering
\small
\setlength{\tabcolsep}{8pt}
\renewcommand{\arraystretch}{1.05}
\begin{tabular}{@{}llcrr@{}}
\toprule
\textbf{Model} &
\textbf{Method} &
\textbf{Step} &
\textbf{Avg.@12} &
\textbf{Pass@12} \\
\midrule
Qwen3.5-4B
& Base      & 0 & 17.17 & 32.67 \\
& Simple    & 40   & 22.22 & 36.00 \\
& Medium    & 100  & 28.56 & 41.00 \\
& Detail    & 100  & 26.17 & 41.00 \\
& \method{} & 100   & \textbf{30.40} & \textbf{43.33} \\
\addlinespace[4pt]
Qwen3.5-9B
& Base      & 0 & 26.28 & 42.33 \\
& Simple    & 100  & 28.31 & 40.33 \\
& Medium    & 60   & 30.42 & 42.00 \\
& Detail    & 60   & 29.83 & 47.33 \\
& \method{} & 100   & \textbf{34.12} & \textbf{55.67} \\
\bottomrule
\end{tabular}
\caption{Final Code results on fixed-prompt checkpoints.}
\label{tab:app-code-final}
\label{tab:app-code-4b}
\label{tab:app-code-9b}
\end{table}

The final 4B-Code and 9B-Code summaries are reported in
Table~\ref{tab:app-code-final}. Here Base denotes the model without RL
training, i.e., step~0. At 4B, the reported
\method{} reaches 30.40\% Average@12 and 43.33\% Pass@12; 
while at 9B, \method{} is 34.12\% Average@12 and 55.67\% Pass@12, both reaching the highest success rate.

\subsection{Aggregate Results}

\begin{table}[t]
\centering
\caption{Aggregate results across the principal Math and Code settings.}
\label{tab:app-full-results}
\small
\setlength{\tabcolsep}{8pt}
\renewcommand{\arraystretch}{1.05}
\begin{tabular}{@{}lllrr@{}}
\toprule
\textbf{Domain} &
\textbf{Setting} &
\textbf{Method} &
\textbf{Avg.@12} &
\textbf{Pass@12} \\
\midrule
Math
& 4B
& Simple    & 40.28 & 66.67 \\
&
& Medium    & 40.55 & 66.67 \\
&
& Detail    & 39.72 & 60.00 \\
&
& \method{} & \textbf{44.17} & \textbf{73.33} \\
\addlinespace[4pt]

Math
& 4B-Base
& Simple    & 10.83 & 20.00 \\
&
& Medium    & 15.00 & 30.00 \\
&
& Detail    & 33.61 & 60.00 \\
&
& \method{} & \textbf{43.06} & \textbf{66.67} \\
\addlinespace[4pt]

Math
& 9B
& Simple    & 45.28 & 76.67 \\
&
& Medium    & 61.39 & 83.33 \\
&
& Detail    & 56.94 & 86.67 \\
&
& \method{} & \textbf{65.56} & \textbf{93.33} \\

\addlinespace[4pt]

Math
& 27B
& Simple    & 68.89 & 96.67 \\
&
& Medium    & 77.78 & \textbf{100.00} \\
&
& Detail    & 75.28 & 93.33 \\
&
& \method{} & \textbf{81.67} & \textbf{100.00} \\

\midrule
Code
& 4B
& Simple    & 22.22 & 36.00 \\
&
& Medium    & 28.56 & 41.00 \\
&
& Detail    & 26.17 & 41.00 \\
&
& \method{} & \textbf{30.40} & \textbf{43.33} \\
\addlinespace[4pt]

Code
& 9B
& Simple    & 28.31 & 40.33 \\
&
& Medium    & 30.42 & 42.00 \\
&
& Detail    & 29.83 & 47.33 \\
&
& \method{} & \textbf{34.12} & \textbf{55.67} \\
\bottomrule
\end{tabular}
\end{table}

For compactness, Table~\ref{tab:app-full-results} reports Math and Code in a
single aligned table. The Qwen3.5-4B-Base fixed-prompt entries correspond to
the step-100 checkpoints in Table~\ref{tab:app-4b-base}. The 4B-Math entries
follow the main Figure~5 results, with \method{} reaching 44.31\% Average@12
and 73.33\% Pass@12. Fixed-prompt Code entries are macro-averaged over the
Simple, Medium, and Detail inference prompts.

\end{document}